\documentclass[letterpaper, 10 pt, conference]{ieeeconf}  % Comment this line out if you need a4paper
\IEEEoverridecommandlockouts % This command is only needed if you want to use the \thanks command
\usepackage{amsmath} % assumes amsmath package installed
\usepackage{amssymb}  % assumes amsmath package installed
\usepackage{color}
\usepackage{graphicx} 
\usepackage{tikz,tikzscale,pgfplots}
\usepackage{epstopdf}
\usepackage{subfigure}
\usepackage{tabularx}
\DeclareMathOperator*{\argminA}{arg\,min}

\DeclareMathOperator*{\logvec}{log\,vec}
\usepackage{url}

\title{\LARGE \bf
Experimental Evaluation of Methods for Estimating Frequency Response Functions of a 6-axes Robot}

\author{Stefanie A. Zimmermann$^{1}$, Stig Moberg$^{2}$
\thanks{$^{1}$Department of Electrical Engineering, Linköping University, Sweden, \tt\small stefanie.zimmermann@liu.se}
\thanks{$^{2}$ABB Robotics, Västerås, Sweden, \tt\small stig.moberg@se.abb.com}}%

\begin{document}

\maketitle
\thispagestyle{empty}
\pagestyle{empty}

%%%%%%%%%%%%%%%%%%%%%%%%%%%%%%%%%%%%%%%%%%%%%%%%%%%%%%%%%%%%%%%%%%%%%%%%%%%%%%%%
\begin{abstract}
Nonparametric estimates of frequency response functions (FRFs) are often suitable for describing the dynamics of a mechanical system. If treating these estimates as measurement inputs, they can be used for parametric identification of, e.g., a gray-box model.
Classical methods for nonparametric FRF estimation of MIMO systems require at least as many ex\-periments as the system has inputs. 
Local parametric FRF estimation methods have been developed for avoiding multiple experiments. 
In this paper, these local methods are adapted and applied for estimating the FRFs of a 6-axes robotic manipulator, which is a nonlinear MIMO system operating in closed loop. The aim is to reduce the experiment time and amount of data needed for identification. The resulting FRFs are analyzed in an experimental study and compared to estimates obtained by classical MIMO techniques. It is furthermore shown that an accurate parametric model identification is possible based on local parametric FRF estimates and that the total experiment time can be significantly reduced.

\end{abstract}

%%%%%%%%%%%%%%%%%%%%%%%%%%%%%%%%%%%%%%%%%%%%%%%%%%%%%%%%%%%%%%%%%%%%%%%%%%%%%%%%
\section{Introduction and related work}

Accurate models of robotic manipulators are needed for many purposes, and they are of particular importance for control design. Today, most control algorithms are based on parametric models, describing the dynamic behavior of the robot in all possible applications.
Combining prior knowledge and a large amount of experimental data, high-fidelity models can be identified \cite{Wernholt_PhD.2007}. The aim of this work is to reduce {the experiment time} while keeping the model's quality. {This will make the identification procedure cheaper}, and will furthermore reduce wear during measuring.

{This paper follows a frequency-domain approach for parameter identification of a nonlinear gray-box model. Linearizations of the system in different configurations are used, while requiring the identified model to be global w.r.t.~configuration, pay- and armloads.}
Aiming to identify a parametric robot model, the first step is the estimation of the system's frequency response function (FRF). 
{This has been done with classical nonparametric methods, which require at least as many experiments as the system has inputs. For improving the quality of the FRF estimate, additional experiments are done and averaging techniques applied. At the cost of many experiments, this method has shown to give accurate FRF estimates, assuming appropriate excitation signals \cite{Wernholt_pC.2008}.}

{In order to reduce the experiment time, this paper evaluates a class of potentially data-efficient methods, local parametric FRF estimation methods, and compare these to the classical nonparametric methods.}
The key idea of local parametric methods \cite{Pintelon.2010.Part_I, Pintelon.2010.Part_II} is that the FRF is a smooth function of the frequency so that it can be approximated in a narrow frequency band around a central frequency by, e.g., a polynomial (Local Polynomial Method, LPM), or a rational function (Local Rational Method, LRM).
{This method allows to estimate the MIMO-FRFs with only one experiment.}

In \cite{Pintelon.2010.Part_I}, the theory is developed for linear dynamic multivariable output error problems. 
In \cite{Pintelon.2010.Part_II}, it is shown that the proposed method can be used for nonlinear systems and the methodology is generalized to handle noisy input-output data (errors-in-variables problem), as well as identification during feedback. 
A modification of the LPM that takes into account constraints between the coefficients of the polynomials at neighboring frequencies was proposed in \cite{Gevers.2011}, allowing reduction of the Mean Square Error of the FRF estimates.
Recently, the local techniques have been revisited \cite{Pintelon.2021} and a new method combining LPM and LRM has been developed, including an automatic local model-order selection procedure.
Besides the data efficiency, the freedom in the multivariable rational model parametrizations is a key aspect of LRM. Different choices for such multivariable rational model parametrizations are discussed in \cite{Voorhoeve.2018}.

In this paper, the LRM is applied for estimating the FRFs of a robotic manipulator, a nonlinear MIMO system operating in closed loop. The final goal is to identify a parametric model based on these FRFs, which is suitable for control purposes. {LRM has been analyzed in the context of Spectral Analysis (e.g.~\cite{Voorhoeve.2018}), while in this paper, }two multivariable LRM parametrizations are experimentally compared with classical, purely nonparametric, averaging techniques that involve multiple experiments. {Furthermore, we propose to combine multiple LRM estimates by logarithmic averaging for improving the estimate.} 
{The main contribution of this paper is an experimental study showing that 
\begin{itemize}
\item LRM estimates allow better parametric gray-box identification than classical nonparametric estimates,
\item for MIMO systems, the choice of parametrization in LRM is crucial and impacts the FRF quality,
\item considering the reference signal improves LRM estimates significantly (Joint Input-Out\-put approach), but classical nonparametric estimates only slightly,
\item logarithmically averaging multiple LRM estimates improves the FRF quality.
\end{itemize}
}
Local parametric and classical nonparametric methods for FRF estimation are reviewed in Section~\ref{sec:FRF_estimation}.
Section~\ref{sec:parameter_estimation} introduces the parametric robot model to be identified and the method for parameter estimation.
Experimental results are presented in Section~\ref{sec:experimental_results}, and conclusions in Section~\ref{sec:conclusions}.

%=======================================================
%  FRF estimation
%=======================================================
\section{Nonparametric FRF estimation}
\label{sec:FRF_estimation}

\begin{figure}[t]
\vspace{2mm}
\begin{center}
\begin{minipage}{.75\columnwidth}
	\def\svgwidth{\columnwidth}
    	\fontsize{7}{7}\selectfont
	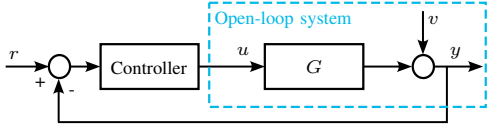
\end{minipage}
\caption{Linear closed-loop system.}
\label{fig:CL_system}
\end{center}
\end{figure}

Consider the setting in Figure~\ref{fig:CL_system}, where $u$ is the plant input (dimension $n_u$) and $y$ is the measured output (dimension $n_y$), corrupted by measurement noise $v$. 
The goal is to obtain a nonparametric estimate of the system's transfer function $G$, as well as the noise covariance matrix {$C_v = \text{Cov}(v)$}.
{The manipulator system is nonlinear with gravity acting as a position-dependent disturbance on the input torque. Without feedback control active, the arm would drift away from the chosen configuration, making the linearization invalid and also introducing a risk for collision. To protect the manipulator and to provide useful data for the identification, experiments are performed in closed loop.} The controller input is the difference between the reference signal $r$ and the output $y$.
The discrete Fourier transforms (DFTs) of the input and output signal are denoted as $U(\omega_k)$ and $Y(\omega_k)$, respectively, 
with the frequencies $\omega_k = k\frac{2\pi}{N T_s} ,\;k=1,2,\dots,N $, 
where $N$ is the total number of samples and $T_s$ the sampling period. 
The DFTs of input and output are related as
\begin{align}
	Y(\omega_k) = G(e^{j\omega_k T_s}) U(\omega_k) + V(\omega_k)
	\label{eq:Y_is_G_times_U}
\end{align}
where $G(e^{j\omega_k T_s})$ is the system's $n_y \times n_u$ transfer function and $V(\omega_k)$ the noise DFT. In the following, the complex variable $\Omega_k = e^{j\omega_k T_s}$ is used.

%=======================================================
%   FRF estimation - Classical purely nonparametric techniques
%=======================================================
\subsection{Classical nonparametric estimation for MIMO systems}
\label{ssec:avg_methods}

Using classical nonparametric techniques for obtaining an estimate of $G(\Omega_k)$, $n_e \geq n_u$ experiments are needed, since \eqref{eq:Y_is_G_times_U} contains $n_y\,\cdot\,n_u$ unknown transfer functions.
The data vectors from $n_e$ experiments are collected into matrices (bold-face) where each column corresponds to one experiment. In the noise-free case, {Eq.~}\eqref{eq:Y_is_G_times_U} becomes
\begin{align}
	\mathbf{Y}(\omega_{k}) = G(\Omega_{k}) \mathbf{U}(\omega_{k})
	\label{eq:DFT_matrix}
\end{align}
where $\mathbf{Y}(\omega_{k})$ and $\mathbf{U}(\omega_{k})$ are $n_y\times n_e$ and $n_u\times n_e$.
{Since the solvability of \eqref{eq:DFT_matrix} depends on the condition number of $\mathbf{U}$, an orthogonal random phase multisine signal is suggested for excitation \cite{Dobrowiecki.2007}.}
Different methods have been proposed for obtaining an estimate $\hat G(\Omega_k)$, e.g.~the H1-estimator \cite{Guillaume.1996}:
\begin{align}
	\hat G^{{\text{H1}}}(\Omega_k) = \mathbf{Y}(\omega_k)  \mathbf{U}^\text{H}(\omega_k) \left[ \mathbf{U}(\omega_k)  \mathbf{U}^\text{H}(\omega_k)\right]^{-1}
\end{align}
{where $(\cdot)^\text{H}$ denotes the conjugate transpose.}

For improving the quality of the estimate, we collect more data and assume that $n_e=M\,n_u$ experiments are done. Then, the DFT matrices can be partitioned into $M$ blocks as
\begin{align}
	\mathbf{X}(\omega_k) &= \left[ \mathbf{X}^{[1]}(\omega_k)  \dots \mathbf{X}^{[M]}(\omega_k)\right]
\label{eq:block_def}
\end{align}
where $\mathbf{X}$ stands for any of the signals $\mathbf{U}$, $\mathbf{Y}$ or $\mathbf{R}$.
Then, the H1-estimator becomes
\begin{align}
\begin{split}
	\hat G^{{\text{H1}}} = 
	\left[\frac{1}{M} \sum_{m=1}^M \mathbf{Y}^{[m]}  {\mathbf{U}^{[m]}}^\text{H} \right] 
	\left[\frac{1}{M} \sum_{m=1}^M \mathbf{U}^{[m]}  {\mathbf{U}^{[m]}}^\text{H} \right]^{-1}
\end{split}
\label{eq:G_H1}
\end{align}
Another method is the arithmetic mean estimator \cite{Guillaume.1998}, given by
$\hat G^\text{ARI} = \frac{1}{M} \sum_{m=1}^M \hat G^{[m]}$, where $\hat G^{[m]} = \mathbf{Y}^{[m]} \left[ \mathbf{U}^{[m]}\right]^{-1}$.
The averaging can be generalized to nonlinear techniques, as in \cite{Guillaume.1998}.
For a robotic system, {and assuming $n_u=n_y$,} the logarithmic technique has shown to give best results \cite{Wernholt_pC.2008}, estimating the FRF by
\begin{align}
	\hat G^\text{LOG} = P^{-1} \text{exp}\left( \frac{1}{M} \sum_{m=1}^M \text{log}\left( P\, \hat G^{[m]} \right) \right)
	\label{eq:G_LOG}
\end{align}
where the matrix $P$ is used to avoid phase wrapping problems when averaging the phase. It is chosen as
\begin{align}
	P = V^{[1]} \text{diag}\left\{ e^{-j \text{arg}\lambda_l^{[1]}} \right\}_{l=1}^n [V^{[1]}]^{-1}
\end{align}
with the eigenvalue decomposition
$ \hat G^{[1]} = V^{[1]} \Lambda^{[1]} [V^{[1]}]^{-1}$, 
$\Lambda^{[1]} = \text{diag}\left\{ \lambda_l^{[1]} \right\}_{l=1}^n$.
{The log- and exp-functions are matrix functions $f(A), A\in \mathbb{C}^{n\times n}$, using the eigenvalue decomposition $f(A) = V f(\Lambda) V^{-1} = V \text{diag}\left\{ f(\lambda_l) \right\}_{l=1}^n V^{-1}$, see \cite{Guillaume.1998}.}
The uncertainty is estimated as \cite{Wernholt_pB.2007}:
\begin{align*}
	\sigma_{\hat{G}}^2 
	= \frac{1}{M^2}\, \sum_{m=1}^M
	\text{vec}(\hat G^{[m]} - \hat G^{{\text{LOG}}})
	\left(\text{vec}(\hat G^{[m]} - \hat G^{{\text{LOG}}})\right)^\text{H}
\end{align*}

Now, the closed-loop system of Figure~\ref{fig:CL_system} is considered and we assume that the reference signal $r$ is available.
Then, a joint input-output (JIO) approach can be applied:
\begin{align}
\hat G^{{\text{JIO}}} =  \left[ \frac{1}{M} \sum_{m=1}^M \mathbf{Y}^{[m]} {\mathbf{R}^{[m]}}^\text{H}\right]
		 	\left[ \frac{1}{M} \sum_{m=1}^M \mathbf{U}^{[m]} {\mathbf{R}^{[m]}}^\text{H}\right]^{-1}
\label{eq:JIO}
\end{align}
This JIO estimator is consistent and asymptotically unbiased and is therefore expected to give the best performance when the number of measured data blocks $M$ increases. This is also concluded in \cite{Wernholt_pB.2007}, where different averaging techniques are compared in a simulation study with a linear robot model.
Compared to that, the experimental results in \cite{Wernholt_pC.2008} show that \eqref{eq:JIO} does not perform that well mainly due to large errors at low frequencies. Since it is based on the H1 estimator, {Eq.~}\eqref{eq:JIO} gives comparable estimates as \eqref{eq:G_H1}, especially if only few experiments are considered, {see Table~\ref{tab:FRF_bias}.}
% 

%=======================================================
%  FRF estimation - LPM
%=======================================================

\subsection{Local polynomial method}
\label{ssec:LPM}

While data from $n_e \geq n_u$ experiments is needed for classical nonparametric FRF estimates, only one {experiment} is sufficient for obtaining an estimate of $G(\Omega_k)$ with local parametric methods.
The key idea of the LPM method is that $G(\Omega_k)$ is a smooth function of the frequency so that it can be approximated in a {local} band around a central frequency by a complex polynomial \cite{Pintelon.2010.Part_I}. 
Assuming a high {signal-to-noise} ratio, the open loop system indicated in Figure\;\ref{fig:CL_system} is considered first. {The closed-loop problem will be considered in Section~\ref{ssec:LRM_JIO}.}
To begin with, every row $i=1,\dots, n_y$ of ${G}$ is estimated separately by considering all $n_u$ inputs ${U}=[U_1, \dots, U_{n_u}]^T$ and one output $Y_i$ {at a time} (MISO).
Then, {Eq.~}\eqref{eq:Y_is_G_times_U} at DFT line $\omega_{k+{\tilde r}}$ can be {approximated} with
\begin{align}
	Y_i(\omega_{k+{\tilde r}}) = {G}_i(\Omega_{k+{\tilde r}}) {U}(\omega_{k+{\tilde r}}) + V_i(\omega_{k+{\tilde r}})
	\label{eq:DFT_1}
\end{align}
where ${G}_i(\Omega_{k+{\tilde r}}) = \left[G_{i1}(\Omega_{k+{\tilde r}}), \dots ,G_{i n_u}(\Omega_{k+{\tilde r}})   \right]$ with
\begin{align}
	&G_{ij}(\Omega_{k+{\tilde r}}) \,=\, 
	G_{ij}(\Omega_{k})+ \sum_{s=1}^R g_{s,ij}(\omega_k) {\tilde r}^s 
	\label{eq:def_G}
\end{align}
for $j=1,\dots,n_u$. 
{Here, $V_i$ denotes both the original noise component and the approximation error.}
$G(\Omega_k)$ is approximated with polynomials of order $R$ in a sliding window of width $w=2 b$, which is centered around a {central frequency $\omega_k$. 
${\tilde r}$ is a component in $r$, where $r = [-b ,\dots,-1,0,1,\dots, b-1]^T$}, except near the frequency borders.
Collecting the data from $w$ {frequency lines} within a window, {Eq.~}\eqref{eq:DFT_1} can be re-written as
\begin{align}
	Y_{w,i} =  {K}_{w,i} {\Theta}_i(\omega_k) + V_{w,i}
	\label{eq:DFT_1b}
\end{align}
where 
$Y_{w,i}$ and $V_{w,i}$ are $w \times 1$ vectors,
and
${\Theta}_i$ is the $(R+1)n_u \times 1$ vector of unknown complex parameters, i.e.:
\begin{align}
\begin{split}
	{\Theta}_i (\omega_k)
	= [ &G_{i1}(\Omega_k) , g_{1,i 1}(\omega_k) , \dots, g_{R,i 1}(\omega_k), \dots, \\ 
	 	&G_{i\,n_u}(\Omega_k) , g_{1,i\,n_u}(\omega_k), \dots ,g_{R,i\,n_u}(\omega_k)]^T
	 	\label{eq:Theta_i_LPM}
\end{split}
\end{align}
${K}_{w,i}$ is the $w \times (R+1) n_u$ matrix containing the input data {$U_w \in \mathbb{C}^{w\times n_u}$} within the window:
\begin{align}
\begin{gathered}
	{K}_{w,i} 
	=  [r^0 \, r^1 \dots r^R] \otimes {U}_w  
	\\= [r^0 U_{w,1} , \dots ,r^R U_{w,1}, \dots , r^0 U_{w,n_u} ,\dots ,r^R U_{w,n_u}] \nonumber
\end{gathered}
\end{align}
{where $\otimes$ means that each column of the $w \times (R+1)$-matrix $[r^0 \, r^1 \dots r^R]$ is multiplied with each column of $U_w$. All products are taken element-wise.}
Note that the window width $w$ must fulfill $w \geq w_{min}=(R+1)n_u$ in order to estimate all parameters ${\Theta}_i$.
If $w > w_{min}$, {Eq.~}\eqref{eq:DFT_1} is an overdetermined set of equations that can be solved using least squares as 
\begin{align}
	{\hat\Theta}_i(\omega_k) = ({K}_{w,i}^\text{H} {K}_{w,i})^{-1} {K}_{w,i}^\text{H} Y_{w,i} 
	\label{eq:lsqr_theta}
\end{align}
For each {central frequency $\omega_k$}, the estimate of the FRF related to output channel $i$ is contained in ${\hat\Theta}_i$ as indicated in \eqref{eq:Theta_i_LPM}.
The residual of the least-squares fit \eqref{eq:lsqr_theta} and an estimate of the noise covariance are given by
\begin{align}
	\hat V_{w,i} = Y_{w,i} - {K}_{w,i} {\hat\Theta}_i(\omega_k) 
	\label{eq:residual}
\\
	\hat C_{V,i}(\omega_k) = \frac{1}{q} \hat V_{w,i}^\text{H} \hat V_{w,i}
	\label{eq:cov}
\end{align}
where $q = w - \text{rank}({K})$, see \cite{Pintelon.2010.Part_I}.

Increasing $w$, i.e.~taking a larger number of frequencies in the frequency window, reduces the variance of the parameter estimate since the noise is averaged over a larger number of data. On the other hand, the larger the window, the larger the  interpolation error caused by the fact that the transfer function varies over the interval.
In practice, the LPM is mostly used with polynomials of degree two, i.e.~$R = 2$ \cite{Gevers.2011}.

%=======================================================
%  FRF estimation - LRM
%=======================================================
\subsection{Local rational method (LRM)}

\subsubsection{MISO parametrization}

Considering rational functions as local approximations, {Eq.~}\eqref{eq:DFT_1} is modified to
\begin{align}
	Y_i(\omega_{k+{\tilde r}}) = \frac{{G}_i(\Omega_{k+{\tilde r}})}{D_i(\omega_{k+{\tilde r}})} {U}(\omega_{k+{\tilde r}}) + V_i(\omega_{k+{\tilde r}})
	\label{eq:DFT_2}
\end{align}
with ${G}_i$ as in \eqref{eq:def_G}. {$D_i$ is mainly added to reduce the ap\-proximation errors around anti-/resonances and is defined as}
\begin{align}
	D_i(\omega_{k+{\tilde r}}) \,= \, 1 + \sum_{s=1}^R d_{s,i}(\omega_k) {\tilde r}^s\,,
	\label{eq:def_D_i}
\end{align}
see Parametrization 2 in \cite{Voorhoeve.2018}.
Multiplying {\eqref{eq:DFT_2}} with $D_i$ gives
\begin{align*}
	&Y_i(\omega_{k+{\tilde r}})  = 
	 \left({{G}_i(\Omega_{k}) }+ \sum_{s=1}^R {g}_{s,i}(\omega_k) {\tilde r}^s \right) {U}(\omega_{k+{\tilde r}}) 		
	\nonumber\\
	&- \left(\sum_{s=1}^R {d}_{s,i}(\omega_k) {\tilde r}^s \right) Y_i(\omega_{k+{\tilde r}}) 		
	+ V_i(\omega_{k+{\tilde r}}) D_i(\omega_{k+{\tilde r}})  
\end{align*}
Using matrix notation yields
\begin{align}
	Y_{w,i}  = {K}_{w,i}  {\Theta}_i(\omega_k) + { \tilde V_{w,i} }
	\label{eq:DFT_2b}
\end{align}
where {$\tilde V_{w,i}$ is the noise term scaled with $D_{w,i}$, which contains \eqref{eq:def_D_i} for all frequencies of the window,}. {${\Theta}_i \in \mathbb{C}^{[(R+1)n_u + R] \times 1}$} is the vector of unknown parameters
\begin{align}
	&{\Theta}_i (\omega_k)
	= [G_{i 1} , g_{1,i 1},  \dots, g_{R,i 1}, \dots, G_{i\,n_u} ,\nonumber\\ 
	&g_{1,i\,n_u}, \dots, g_{R,i\,n_u},  d_{1, i },  \dots, d_{R, i } ]^T  
	\label{eq:theta_i_LRM_MISO}
\end{align}
and {${K}_{w,i} \in \mathbb{C}^{w \times [(R+1) n_u + R]}$} is:
\begin{align}
	{K}_{w,i} = 
	\left[ {[r^0 \, r^1 \dots r^R]} \otimes {U}_w\;,\;
	-[r \dots r^R] \otimes Y_{w,i}\right]
\end{align}
The window width must be $w \geq w_{min}=(R+1)n_u + R$ in order to estimate all parameters ${\Theta}_i$.
\eqref{eq:DFT_2b} is linear in the parameters and can be solved with \eqref{eq:lsqr_theta}, which is referred to as Local Levy method in \cite{Pintelon.2021}. It is noted, but not further considered in the scope of this paper, that the linearization of the output error yields a biased estimate \cite[p. 301 f.]{Pintelon.2012}.
Note also that a transient term is not estimated, since we pre-process the data and use only steady-state samples.

%=======================================================
\subsubsection{MIMO parametrization}
\label{ssec:LRM_MIMO}

Since the robot operates in closed loop, all outputs affect the system through feedback, and we therefore consider {all outputs $Y_i$} simultaneously in the following. Then, ${Y} = [Y_1,\dots,Y_{n_y}]$, and ${D}_w$ becomes a $n_y\times n_y$-matrix of polynomials (full MFD parametrization, see Parametrization 3 in \cite{Voorhoeve.2018}).
Then, analogous to \eqref{eq:DFT_2b},
\begin{align}
	{Y}_w  = {K}_w  {\Theta}(\omega_k) + { \tilde V_w }
	\label{eq:DFT_3}
\end{align}
where ${Y}_w$ and {$\tilde V_w$} are $w \times n_y$ matrices and
${\Theta}(\omega_k)= [{\Theta}_1(\omega_k) , \dots, {\Theta}_{n_y}(\omega_k)] $ 
is the $(R+1)n_u + R\,n_y \times n_y$ matrix of unknown parameters. For $i=1\dots n_y$,
\begin{align}
	&{\Theta}_i(\omega_k) = 
	[G_{i 1} ,\, g_{1,i 1},  \dots, g_{R,i 1}, \dots, G_{i\,n_u} ,\, g_{1,i\,n_u}, \dots,  \nonumber\\ 
	&g_{R,i\,n_u},\; d_{1,i 1},  \dots, d_{R,i 1},  \dots,  d_{1,i\,n_y},  \dots, d_{R,i\,n_y} ]^T 
\end{align}
Now, $w_{min}=(R+1)n_u + R\,n_y$, and 
{${K}_w \in \mathbb{C}^{w \times [(R+1) n_u + R\,n_y]}$} is:
\begin{align}
	{K}_w = 
	\left[  	{[r^0 \, r^1 \dots r^R]} \otimes {U}_w\;,\;
	-[r \dots r^R] \otimes {Y}_w \right]
\end{align}
This approach has the advantage that the cross-influence of the different outputs is taken into account through the parameters of the polynomials in the matrix ${D}$.

Analogous as in \eqref{eq:residual} and \eqref{eq:cov}, the uncertainty of the FRF estimate is derived from the residuals as
{
\begin{align}
{\hat{\tilde{V}}_w}(\omega_k) &= {Y}_w - {K}_w \hat{{\Theta}}(\omega_k) \\
	\hat C_{{\tilde V_w}}(\omega_k) &= \frac{1}{q} {\hat{\tilde{V}}_w}^\text{H}(\omega_k) {\hat{\tilde{V}}_w}(\omega_k) 	 \,.
	\label{eq:FRF_uncertainty_LRM}
\end{align}
}
\vspace{-.3cm}
% ===========================================================
\subsubsection{Joint input-output approach}
\label{ssec:LRM_JIO}

Assuming the closed-loop system of Figure~\ref{fig:CL_system}, and that the reference signal $r$ is available, the following joint input-output (JIO) approach can be applied \cite{Wellstead.1981}, \cite[p. 438]{Ljung.1999}: First, the FRFs $\hat G_{ru}$ and $\hat G_{ry}$ from the reference $r$ to the input $u$, and from the reference $r$ to the output $y$ are estimated using LRM. Second, the FRF from $u$ to $y$ is computed as the ratio $\hat G_{ry}/\hat G_{ru}$. We will call this variant JIO-LRM in the following.

% ===========================================================
\subsubsection{Involving multiple experiments}

For improving the LRM estimate, multiple experiments can be done and logarithmic averaging can be applied. Then, $\hat G^{[m]}$ in \eqref{eq:G_LOG} is replaced by the LRM estimate and $M$ is replaced by $n_e$, since LRM delivers an estimate for each single experiment.

%=======================================================
%  Parametric model estimation
%=======================================================
\section{Parametric model identification}
\label{sec:parameter_estimation}

{The following section is dedicated to the second step of identification, i.e.~the parameter estimation based on a nonparametric FRF estimate. The goal is to fit a parametric gray-box model, which can, e.g., be used for control purposes.}

\subsection{Gray-box model structure}
\label{ssec:graybox_model}

\begin{figure}[bt]
\vspace{2mm}{
\begin{center}
\begin{minipage}{0.8\columnwidth}
	\def\svgwidth{\columnwidth}
    	\fontsize{7}{7}\selectfont
	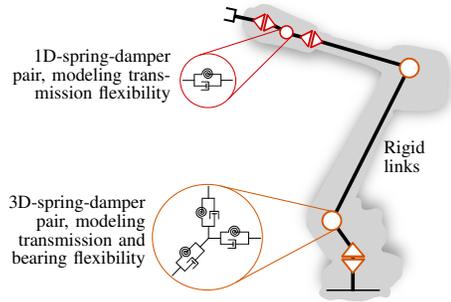
\end{minipage}
\end{center}
\vspace{-2mm}
\caption{Gray-box model of an 6-axes manipulator.}
\label{fig:GB_Model}
}
\end{figure}

{The robot under consideration is a 6-axes serial-link manipulator}. The model structure proposed in \cite{Ohr.2006} is used, where a rigid body model with 6 degrees of freedom is extended by flexibility and friction in the joints. 
The transmission stiffness of joints 4 to 6 is modeled by linear spring-damper pairs, acting in the direction of rotation of the joint.
Due to high loading, the transmission stiffness of joints 1 to 3 is described as nonlinear function, and two more linear spring-damper pairs are added for taking into account bearing and structural flexibility. Figure~\ref{fig:GB_Model} shows a drawing of the robot model.
The vector of joint angles is named $q_a$ or $q_m$ depending on if it is expressed on arm or motor side of the gearbox. A realization of $q_a$ is called \textit{robot configuration}.
The angular motion between the rigid bodies due to elastic effects that act perpendicular to the direction of transmission is described by the variables $q_e$.
{In order to derive a state-space model, we choose} the state vector $x = [q_m, q_a, q_e,   \dot q_m, \dot q_a, \dot q_e]^T$, the applied motor torque as input $u$, and the motor angular velocity $\dot q_m$ as output. {We obtain the equations of motions}:
{
\begin{align}
\begin{split}
\dot x &=f(x,u,\theta) = \begin{bmatrix} 
\dot q_m\\ 
\dot q_a\\ 
\dot q_e\\
M_m^{-1} (u-\tau_{fm}-r_g\tau_g)\\
M_{ae}^{-1}  \left(\begin{bmatrix}\tau_g(\theta)\\\tau_e(\theta)\end{bmatrix} -c_{ae} - g_{ae} \right)
\end{bmatrix}
\\
y &= h(x,u,\theta) = \dot q_m 
\end{split}
\label{eq:ss_model}
\end{align}
where 
$M_m$ is the matrix of motor inertias,
$\tau_{fm}(\dot q_m)$ is the joint friction (assumed on the input side of the gearbox),
$\tau_g$ is the gear torque,
$r_g$ is the matrix of inverse gear ratios,
$M_{ae}(q_a,q_e)$ is the inertia matrix,
$c_{ae}(q_a,q_e,\dot q_a,\dot q_e)$ is the velocity dependent torque,
$g_{ae}(q_a,q_e)$ is the gravity torque,
$k_g$, $k_e$, $d_g$ and $d_e$ are the joint stiffness and damping constants in the direction of transmission (index $g$) and perpendicular to it (index $e$).
}
In the scope of this work, the rigid body parameters are assumed to be known, while a friction model is identified separately. The challenge of identification is therefore restricted to estimating the stiffness and damping parameters collected in $\theta$. 

 \subsection{Frequency domain identification algorithm}
\label{ssec:Freq_id_algorithm}

A frequency-domain method is used for finding $\theta$, as proposed in \cite{Wernholt_pA.2008}. The method is based on the assumption that the excitation signal is a small perturbation around a robot {configuration}, allowing linearization of \eqref{eq:ss_model}.
The parameters $\hat\theta$ are obtained by mi\-nimizing the weighted log-error between the FRFs $\hat G^{(i)}(\omega)$, estimated from measurements (see Section~\ref{sec:FRF_estimation}), and the parametric FRFs $G^{(i)}(\omega,\theta)$ of the linearized gray-box model:
	\begin{align}%
	\begin{split}
		&\hat\theta = \argminA_\theta 
		\sum_{i\in Q_c} \sum_{l=1}^{N_f} 
		\left[\mathcal{E}^{(i)}(\omega_l,\theta)\right]^\text{H}  	
		W^{(i)}(\omega_l) \;  
		\mathcal{E}^{(i)}(\omega_l,\theta)
		\label{eq:cost_function}
		\\
		&\mathcal{E}^{(i)}(\omega_l,\theta) = \logvec(\hat G^{(i)}(\omega_l))  - \logvec(G^{(i)}(\omega_l,\theta))
	\end{split}%
	\end{align}%
	where {$\text{log}(G)=\text{log}|G|+j\,\text{arg}(G)$},
	$N_f$ is the number of frequencies, 
	$Q_c$ a set of robot configurations, and 
	$W^{(i)}(\omega)$ a {weighting matrix. The weights reflect both the uncertainty in $\hat G$, as well as the user choice of which FRF elements and frequency ranges are considered as most important to describe with the model \cite{Moberg.2014}. Here, the diagonal elements of the FRFs and the frequency range around the first antiresonance and resonance are weighted highest.}
{The log-criterion is used because of the large dynamic range of the highly resonant robot system and because of its robustness w.r.t.~outliers in the data \cite{Pintelon.2012}.}
{The problem \eqref{eq:cost_function} is solved using the MATLAB-function \textit{fminunc}. Since the objective function is not convex, reasonable initial values are needed, which can, e.g.,~be found by the approach from~\cite{Zimmermann.2020}. Furthermore, the problem is solved for a number of random perturbations around the initial guess. 
The linearization $G^{(i)}(\omega,\theta)$ is obtained from first order Taylor series expansion of~\eqref{eq:ss_model}. The linearization points $i$ are chosen by the experiment design approach as in~\cite{Zimmermann.2023}. }

%=======================================================
%  EXPERIMENTAL RESULTS
%=======================================================
\section{Experimental results}
\label{sec:experimental_results}

\subsection{Excitation design and measurement signals}
\label{ssec:experimental_setup}

An orthogonal random phase multi-sine signal is chosen for excitation, containing {336} log-spaced odd frequencies in the range 4\,-\,80\,Hz. All motors are excited simultaneously {with a speed reference}.
In order to reduce the effect of static friction, zero velocity is avoided by adding a low-frequent single sine with sufficiently large amplitude to the multi-sine reference \cite{Wernholt_pD.2008}. {Motor torques ($u$) and accelerations ($y$) are logged at 2\,kHz.}
The period time for one experiment is {28\,s}.
The FRFs are estimated in 49 linearization points, i.e.~49 random robot configurations distributed in the work-space.

Both real experiments and simulations are done. In simulation, a robot model \eqref{eq:ss_model} {with parameters $\theta_0$ is excited around configurations $i$ and its FRFs $G^{(i)}(\omega,\theta_0)$ are computed.} The robot plant is simulated with a controller, 
{consisting of a basic P-type position controller with an inner velocity loop of PI-type. 
White noise and deterministic position dependent disturbances are added to the simulated motor position.  Moreover, deterministic position dependent disturbances are added to the motor torque for simulating motor torque ripple. 
All deterministic disturbances are periodic in the motor position and therefore described as a sum of sinusoids. All disturbances are tuned such that the simulation generates data that is very similar to the real data.}

\subsection{FRF estimation}
\label{ssec:results_FRF_estimation}

The logarithmic FRF estimate {\eqref{eq:G_LOG}} with $M=4$ blocks of data serves as reference, i.e.~data from $n_e=M n_u = 24$ experiments is used. 
Estimates involving fewer experiments are compared.
$\hat G$ is derived with the $\text{(MISO-, MIMO-, JIO-)}$ LRM and H1 methods using only $n_e=6$ experiments. Note that the H1 and LOG estimator are identical if $M=1$, i.e.~$n_e=6$. The MIMO parametrization of Section~\ref{ssec:LRM_MIMO} is used for JIO-LRM, and logarithmic averaging is applied for merging the 6 LRM estimates. 
Furthermore, an estimate using JIO-LRM that is derived from only one experiment is included.
For the LRM, second order polynomials are used, i.e.~$R=2$, and a window slightly larger than $w_{min}$ is chosen.

Figure~\ref{fig:FRF_comp_diag} shows the FRF estimates of an exemplary robot configuration obtained from real data. 
{The H1 method \eqref{eq:G_H1} gives a noisy estimate if only 6 experiments are used, which is the minimum for this method. Collecting more data by exciting the motors with different realizations of the reference signal, logarithmic averaging can be applied. Figure~\ref{fig:FRF_comp_diag} shows the LOG estimate \eqref{eq:G_LOG} with $M=4$ blocks data.} Note that no windowing technique is applied, i.e.~the FRF is computed at every excited frequency separately and averaged with the estimate from other experiments, which are independent realizations of the input signal.
{Assuming the LOG estimate as \textit{the truth} and comparing the FRFs estimated with LRM,} the most obvious observation is that the damping of the resonances is overestimated using MIMO-LRM. Compared to that, the anti-resonances can be estimated more accurately. The fact that the resonance peaks are hard to estimate with LRM has been observed in \cite{Pintelon.2021}, and has been explained by overemphasis of the noise at the borders of the local frequency band.
The JIO-LRM estimate is much better around the resonance peaks, indicating that the local method is more sensitive to feedback effects compared to classical averaging techniques. {Even the JIO-LRM obtained from only one experiment succeeds in estimating the frequency location of the resonances. The damping is much lower compared to the LOG-estimate and the estimate for low frequencies yields a large error for the robot's wrist axes ($G_{44}$ to $G_{66}$).}

\begin{figure}[tb]
\vspace{-4mm}
\begin{minipage}{1.16\columnwidth}
	\hspace{-.8cm}
	\fontsize{8}{8}\selectfont
	\includegraphics[width=\columnwidth]{FRF_real_6x6.tikz}
\end{minipage} 
\caption{Diagonal elements of $\hat G$ estimated from $n_e$ experiments ($n_e$ indicated as indices in the legend).}
\label{fig:FRF_comp_diag}

\vspace{2mm}

\begin{minipage}{\columnwidth}
	\fontsize{8}{8}\selectfont
	\includegraphics[width=\columnwidth]{FRF_real_G33.tikz}
\end{minipage} 
\caption{Enlargement of $G_{3,3}$.}
\label{fig:FRF_33}
\end{figure}

% ================================================

For further validation of the different FRF estimates, {a nonlinear robot model with parameters $\theta_0$ is simulated in closed loop with added disturbances.} A weighted amplitude bias is computed as
{
\begin{align}
	\frac{1}{N_f\, Q} \sum_{i=1}^{Q}
	\sum_{l=1}^{N_f} 
	\left\lVert W^T(\omega_l)  \left[|G^{(i)}(\omega_l,\theta_0)|- |\hat G^{(i)}(\omega_l)|\right] \right\rVert_2
	 \label{eq:FRF_cost}
\end{align}
}%
where $Q$ is the number of robot configurations and $W$ a weighting matrix{, $\|W(\omega_l)\|=1$}, favoring the diagonal entries of $G$ and $\hat G$. {$|\cdot|$ denotes the elementwise absolute value and $\|\cdot\|_2$ the matrix 2-norm.}
{Note that the bias \eqref{eq:FRF_cost} will be greater than zero due to, e.g., deficiencies in the data, nonlinearities and the controller impact.} Table~\ref{tab:FRF_bias} shows the bias for $Q=49$ random configurations.
The MIMO-LRM estimate gives a smaller bias compared to MISO-LRM, if $n_e > 1$, demonstrating that the choice of parametrization is crucial for the method. Since excitation of one axis affects all other axes, the outputs cannot be treated independently, which needs to be considered in the polynomials of $D$ in \eqref{eq:DFT_3}. Using LRM, the JIO-approach leads to a significant bias reduction. {Compared to that, the JIO estimate \eqref{eq:JIO} yields only a small improvement compared to LOG \eqref{eq:G_LOG}.
In order to achieve short measurement times, minimal estimation data shall be used, i.e.~the last entries in each column of Table~\ref{tab:FRF_bias} are compared: JIO-LRM with $n_e=1$ gives the lowest FRF amplitude bias. In particular, it is lower than the bias of averaged estimates with 1 block of experiments ($n_e=6$).}

\begin{table}[t]
\vspace{2mm}
\begin{center}
\caption{Bias of FRF amplitude \eqref{eq:FRF_cost}; Avg.~of 49 robot configurations.}
\label{tab:FRF_bias}
\setlength\extrarowheight{1pt}
\begin{tabularx}{\columnwidth}{|X ||p{.87cm}|p{.87cm}||p{.87cm}|p{.87cm}|p{.87cm}|}
\hline
	$n_e$ per configuration	  & LOG (Eq.~\eqref{eq:G_LOG}) 	& JIO (Eq.~\eqref{eq:JIO}) 	& LRM, MISO 	& LRM, MIMO & LRM, JIO 	\\ \hline\hline
	24 ($M=4$)	& 135.55	& 134.72 & 248.65 & 240.40		& 125.22	 	\\    \hline
  	6 ($M=1$)	& 162.14 & 162.14 & 257.71	& 249.89		& 136.20	 	\\    \hline
  	1			& \textendash\ & \textendash\ 	& 238.89 & 238.90		& 147.27	 	\\    \hline
\end{tabularx}
\end{center}
\end{table}

\subsection{Parametric model identification}

{It was shown in the previous section that additional experiments, i.e.~$n_e>1$, improve the FRF accuracy.} Aiming to estimate the parameters of the gray-box model \eqref{eq:ss_model} with data collected in minimal experiment time, a trade-off needs to be made between experiment time and FRF quality. 
In this section, the stiffness parameters of the model \eqref{eq:ss_model} are estimated using the algorithm in Section~\ref{ssec:Freq_id_algorithm}. The different nonparametric FRF estimates from Section \ref{ssec:results_FRF_estimation} are used for estimation and the resulting model parameters $\theta$ are compared. {The aim is to find a FRF estimation method suitable in the context of parametric gray-box identification.}

In order to measure the quality of the derived parametric models, data is obtained from simulation where the true model with parameters $\theta_0$ is known.
{The JIO-LRM and the LOG\footnote{Since JIO (Eq.~\eqref{eq:JIO}) gives a similar FRF bias (Table~\ref{tab:FRF_bias}) but requires the recording of the reference signals, LOG (Eq.~\eqref{eq:G_LOG}) is preferred.} methods} are compared in Table~\ref{tab:bias_sim} {w.r.t.~their average bias $\bar B_\theta = \frac{1}{p}\sum_{s=1}^{p} \frac{|\theta_{0,s}-\hat\theta_s|}{\theta_{0,s}} $, $p=\text{dim}(\theta)$}. The estimation data was generated in {7 robot configurations, found by improved experiment design as proposed in \cite{Zimmermann.2023}.} Either 24, 6 or 1 experiments were simulated in each configuration.
{A bias occurs between $\theta_0$ and $\hat \theta$ due to unmodeled nonlinearities such as nonlinear transmission stiffness, and closed-loop effects. The bias of most parameters in $\theta$ is less than 3\,\%, but two parameters have a bias in the range of 20-50\,\%.}

The average parameter bias is lowest if data from 24 experiments is used together with the {JIO-LRM method. In general, the JIO-LRM outperforms the traditional LOG-avg method, if the same experiment time is assumed. Note that JIO-LRM requires the reference signals to be recorded.
Even though the FRF amplitude bias for JIO-LRM with $n_e=1$ is lower than the bias of LOG with $n_e=6$ (Table~\ref{tab:FRF_bias}), the parametric model which is estimated from the latter is more accurate in terms of $\bar B_\theta$. 
It should also be noted that the results in Table\;\ref{tab:bias_sim} depend on the chosen excitation signal, as well as the robot configurations selected for the experiments.
Depending on the intended use of the parametric model, a larger bias in some components of $\theta$ might be acceptable.
From a practical point of view, a trade-off needs to be made between parameter accuracy and experiment time.}

\begin{table}[tb]
\vspace{2mm}
\centering \scriptsize
\caption{Mean bias {$\bar B_\theta = \frac{1}{p}\sum_{s=1}^{p} \frac{|\theta_{0,s}-\hat\theta_s|}{\theta_{0,s}} $} of the model parameters; Estimation data from $n_e$ experiments in 7 configurations.}
\label{tab:bias_sim}
\setlength\extrarowheight{1pt}
\begin{tabularx}{\columnwidth}{|X || c || c|c || c|c |}
    \hline
		& \multicolumn{1}{c||}{$n_e=1$} 	& \multicolumn{2}{c||}{$n_e=6$}  	& \multicolumn{2}{c|}{$n_e=24$}   
	\\ 	\hline
	Method & JIO-LRM & LOG \eqref{eq:G_LOG} 	& JIO-LRM   	& LOG \eqref{eq:G_LOG} 	& JIO-LRM  
	\\ 	\hline\hline
	{$\bar B_\theta$} & 19.1\,\%	 & 10.0\,\% 	& 9.1\,\%  & 7.9\,\% 	& 4.8\,\%
	\\ 	\hline 
\end{tabularx}
\end{table}

%=======================================================
%  CONCLUSION
%=======================================================
\section{Conclusions}
\label{sec:conclusions}

In order to keep the experiment time as low as possible, the data-efficient LRM method was applied for estimating nonparametric FRFs of a 6-axes robot. The resulting FRFs were compared to {classical, purely nonparametric, estimates obtained from multiple experiments. Furthermore logarithmic averaging of multiple FRF estimates was compared.}
The presented results of an experimental study showed that a MIMO parametrization in LRM gives a more accurate FRF estimate compared to a MISO parametrization. Furthermore, the benefit of a JIO approach combined with LRM was shown, {as well as the possible improvement by additional experiments and logarithmic averaging}.
The presented results suggest that the {JIO-LRM method combined with} LOG-averaging gives the most accurate FRF estimate among the compared methods, assuming the same measurement time. 
Since the LRM estimate with minimum data, i.e.~one experiment, {gives only poor identification results} for the parametric robot model, it is concluded that a trade-off needs to be made between experiment time and identification accuracy. 
Further research on local parameterizations as well as the effects of nonlinearities in the system is recommended in order to improve the LRM estimate and in order to reduce the parameter bias of the gray-box robot model.

\section{Acknowledgments}
This work was sponsored by the Vinnova competence center LINK-SIC.

\addtolength{\textheight}{-15cm}

\end{document}